\documentclass{article}

\usepackage{arxiv}

\usepackage[utf8]{inputenc}
\usepackage[T1]{fontenc}
\usepackage{amsmath,amssymb}
\usepackage{graphicx}
\usepackage{multirow}
\usepackage{booktabs}
\usepackage{url}
\usepackage[hidelinks]{hyperref}

\renewcommand{\headeright}{Preprint}
\renewcommand{\undertitle}{Preprint}
\renewcommand{\shorttitle}{Vision and Text Transformer for Predicting Answerability on VQA}

\title{Vision and Text Transformer for Predicting Answerability\\ on Visual Question Answering}

\author{
  Tung Le\thanks{Japan Advanced Institute of Science and Technology, Ishikawa, Japan.} \,\thanks{Faculty of Information Technology, University of Science, Ho Chi Minh City, Vietnam; Vietnam National University, Ho Chi Minh City, Vietnam.} \\
  \texttt{lttung@jaist.ac.jp} \\
  \And
  Huy Tien Nguyen\footnotemark[2] \,\thanks{Zalo Research Center, Ho Chi Minh City, Vietnam.} \\
  \texttt{ntienhuy@fit.hcmus.edu.vn} \\
  \And
  Minh Le Nguyen\footnotemark[1] \\
  \texttt{nguyenml@jaist.ac.jp} \\
}

\begin{document}
\maketitle

\begin{abstract}
Answerability on Visual Question Answering is a novel and attractive task to predict answerable scores between images and questions in multi-modal data. Existing works often utilize a binary mapping from visual question answering systems into Answerability. It does not reflect the essence of this problem. Together with our consideration of Answerability in a regression task, we propose VT-Transformer, which exploits visual and textual features through Transformer architecture. Experimental results on VizWiz 2020 dataset show the effectiveness and robustness of VT-Transformer for  Answerability on Visual Question Answering when comparing with competitive baselines.
\end{abstract}

\keywords{Answerability \and Visual Question Answering \and VizWiz \and Vision Transformer \and Multi-head Attention}

\section{Introduction}
\label{sec:intro}
In the rapid increasing of multi-modal information like videos, images, and texts, it has arisen a trend of interdisciplinary studies in Vision and Language. The goals of those researches are to reveal the relationship between visual and textual information to support multi-modal tasks such as Visual Question Answering~\cite{vqa_icip2020, bertrg}, Visual Commonsense Reasoning~\cite{Wang_2020_CVPR}, Image Captioning~\cite{Yun_2019_ICCV} and so on.  

In this work, we focus on an attractive and novel task, Answerability on Visual Question Answering (VQA), which is recently proposed in a contest of VQA for blind people~\cite{dataset2018,dataset2019} in 2018. Generally, Answerability is to determine a score that reflects the answerable ability of samples. At first glance, we wonder whether this task is well worth our researches or not. It is useful and practical enough to help us in deciding to answer a question from users. If we can regard a sample as an unanswerable one, we can quickly respond to users instead of running a completed VQA model. It is beneficial to save computational cost in question answering phases.

\begin{figure}[ht!]
\centering
\includegraphics[width=0.55\textwidth]{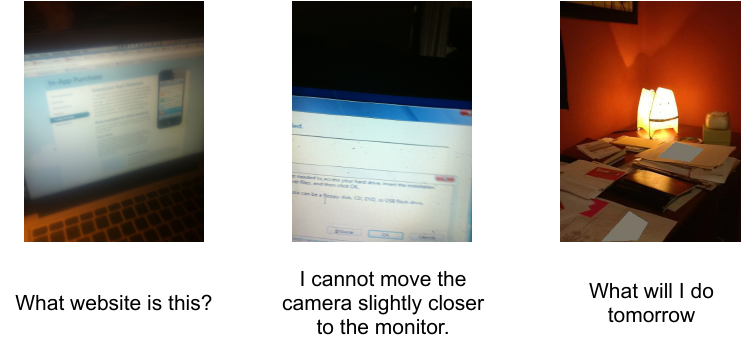}
\caption{Examples of unanswerable sample in VizWiz dataset 2020 (Answerability score = 0)}
\label{fig:example}
\end{figure}

Answerability belongs to multi-modal tasks for understanding hidden features between image and text. In traditional approaches, this task is often not considered as an individual task. Recent results in Answerability are derived from corresponding VQA systems. Firstly, a pre-defined answer's vocabulary of VQA is appended by an ``unanswerable'' label. Then, a pre-defined binary mapping is to convert a predicted VQA label into an answerability score. Specifically, if the VQA model predicts a sample as an answerable one, its answerability score is 1, and otherwise. In those problem statements, Answerability is considered in a classification task. A fixed score of answerability in VQA systems is too far from the problem's essence. Therefore, we propose to consider Answerability in a regression model instead of a binary mapping from VQA. Our proposal comes from two following reasons: (i) evaluation metrics (ii) optimization goal. Firstly, the popular evaluation metric in Answerability is an average precision (AP) scores determined by thresholds in a precision-recall curve. Secondly, VQA systems put their effort into predicting a suitable answer through answer's distribution. However, Answerability reflects problems in samples and annotations. Typical examples of unanswerable samples are presented in Figure~\ref{fig:example} such as poor-quality images, ambiguous questions, and inconsistent data of annotators. The fundamental challenges in Answerability are to extract visual-textual features and their relationship to determine answerability scores. Accordingly, our problem statement, regression Answerability, is not only novel but also powerful enough to overcome Answerability challenges.

Lately, BERT proves significant successes of Transformer~\cite{transformer} architectures in Natural Language Processing (NLP). Accordingly, recent researches in Computer Vision (CV) also gain much more interest in Transformer architectures and their components~\cite{vt, imgTransformer}. Vision Transformer~\cite{vt} is firstly introduced in an image recognition task when integrating Transformer architecture with the fewest mirror adjustments~\cite{vt} into Image Embedding. Specifically, an image is divided into many regions called patches which are put into a Transformer architecture as an Image Encoder. Image patches play similar roles as tokens of sentences in NLP. Obviously, Transformer is useful enough to extract hidden features in images and texts. Advantages of Transformer in both CV and NLP inspire us to integrate them into our Answerability model. 

Despite successes of Transformer in CV and NLP, an important question is how it works in the Answerability problem. Therefore, we propose a regression model combining the strength of Transformer in Text and Vision to overcome challenges in Answerability. Furthermore, we also take advantage of pre-trained models in our architecture to enhance its performance and robustness. In experiments, our models outperform previous baseline approaches in the VizWiz-2020 dataset. Our main contributions are as follows: (i) We introduce a novel problem statement in Answerability, which firstly introduces in the research. (ii) We propose a Vision-Text Transformer model in the Answerability task. (iii) Our proposed model proves its performance and robustness in a practical and novel dataset, VizWiz 2020.

\section{Methodology}
\label{sec:method}
\subsection{Question embedding}
Recently, a proposal of BERT~\cite{devlin-etal-2019-bert} marked a significant moment in linguistic representation.  Particularly, BERT~\cite{devlin-etal-2019-bert} can capture the meaning of words and sentences from their context. It allows one word to have multiple linguistic senses, which is unavailable in Glove~\cite{glove}, Word2Vec~\cite{word2vec} and so on. 

\begin{figure}[ht!]
\centering
\includegraphics[width=0.62\textwidth]{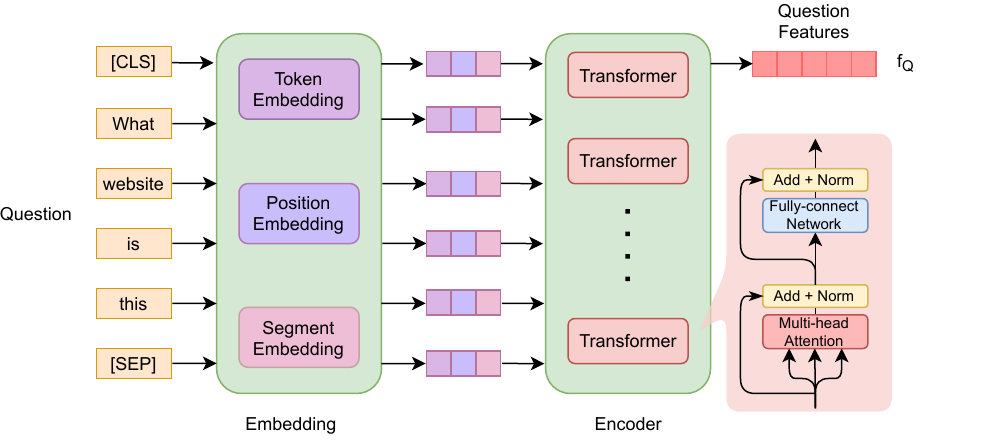}
\caption{Question Embedding: extracts textual features of words in question by BERT - Text Transformer.}
\label{fig:qstEmbedding}
\end{figure}

Specifically, in our architecture, a question is embedded by a pre-trained BERT model. The detail of our question embedding is shown in Figure~\ref{fig:qstEmbedding}. Each question is expanded by two special characters including $[CLS]$ and $[SEP]$ to mark the start and end of the sequence. A question representation is derived from a value of $[CLS]$ vector. 

\subsection{Image embedding}

In traditional approaches, people often process an image through Convolution Neural Networks. However, the successes of Transformer in NLP inspire researches to integrate it into vision area~\cite{odTransformer, wangTransformer}. Among previous approaches, Vision Transformer proves its efficiency in many image classification datasets. However, in our work, Vision Transformer is firstly combined with BERT~\cite{devlin-etal-2019-bert} as Text Transformer.   
\begin{figure}[ht!]
\centering
\includegraphics[width=0.62\textwidth]{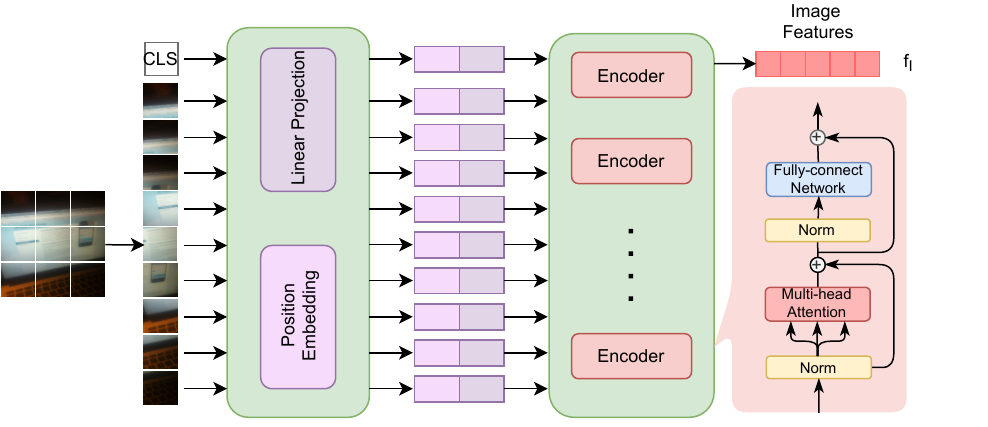}
\caption{Image Embedding: splits an image into patches and convert it via Linear and Positional Embedding to extract the regional and visual features.}
\label{fig:imgEmbedding}
\end{figure}

Specifically, in Vision Transformer, a 2D image is split into a sequence of flattening regions called patches. The number of patches depends on the size of an input image and a predefined patch's size. After, patches are connected with position embedding to retain positional information. Accordingly, Vision Transformer treats an image as a sentence whose tokens are similar to visual regions. In our model, we integrate a pre-trained Vision Transformer to extract visual features of a query image. The detail of our image embedding is presented in Figure~\ref{fig:imgEmbedding}. Specifically, we change the original classifier in Vision Transformer into a fully-connected layer to extract visual features.  

\subsection{Vision-Text transformer answerability model}

After extract visual and textual features in previous modules, we propose a framework to predict an answerability score. 
\begin{figure}[ht!]
\centering
\includegraphics[width=0.58\textwidth]{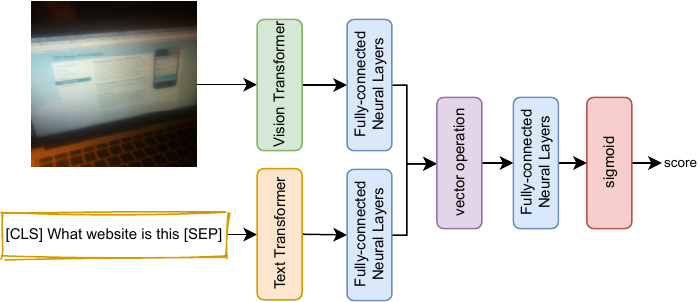}
\caption{VT-Transformer: combines Vision-Text Transformer features by a vector operation to predict an answerable score in the regression model}
\label{fig:model}
\end{figure}
In this architecture, image and question features are normalized by a fully-connected layer into the same dimension space in Equation~\ref{eq:mapping}. 
\begin{equation}
\label{eq:mapping}
   f^{'}_I = W_I^Tf_I + b_I;
   f^{'}_Q = W_Q^Tf_Q + b_Q
\end{equation}
After that, we use a vector operation $\odot$ that includes either multiplication or concatenation to combine them into a meaningful representation. Finally, the answerability score is determined by a fully-connected layer and sigmoid function via Equation~\ref{eq:vecOp}. 
\begin{equation}
\label{eq:vecOp}
   \hat{s_i} = \sigma(W_r^T\left(f^{'}_I \odot f^{'}_Q\right) + b_r) 
\end{equation}
As we mentioned above, the most successful factor of this work is the novel problem statement. Specifically, we propose to consider this task in regression. Therefore, we use Mean Squared Error in Equation~\ref{eq:lossFunc} as our loss function instead of cross-entropy in VQA.
\begin{equation}
\label{eq:lossFunc}
    Loss = \frac{1}{N}{\sum_{i=1}^{n}\left(s_i - \hat{s_i}\right)^2} 
\end{equation}
Where $s_i = \{0,1\}$ corresponds to unanswerable and answerable sample. 
\section{Experiment}
\label{sec:experiment}
\subsection{Dataset and experimental settings}
\subsubsection{VizWiz dataset}
VizWiz dataset~\cite{dataset2018, dataset2019} is published in 2018 and updated until now. It was the first dataset that mentioned the Answerability problem in Visual Question Answering. In the VizWiz Competition, Answerability is regarded as an individual and novel task. In our work, all experiments are conducted and evaluated on the VizWiz dataset. The detail of this dataset is presented in Table~\ref{tab:dataset}. The distribution of unanswerable samples in VizWiz is quite high by 27\% in train and 32\% in the validation set. Respectively, an Answerability system is ideal and essential enough to support VQA models for filtering unanswerable samples that are too hard to predict an answer. 
\begin{table}[ht!]
\centering
\caption{Detail of VizWiz 2020 dataset}
\label{tab:dataset}
\begin{tabular}{lccc}
\hline
\multicolumn{1}{c}{\textbf{Dataset}} & \textbf{Train}                               & \multicolumn{1}{l}{\textbf{Validation}} & \multicolumn{1}{l}{\textbf{Test}} \\ \hline
\textbf{No. Samples}                 & 20523                                                 & 4319                                                & 8000                              \\ \hline
\textbf{\%Answerable}                & \begin{tabular}[c]{@{}c@{}}14981 \\ 73\%\end{tabular} & \begin{tabular}[c]{@{}c@{}}2937\\ 68\%\end{tabular} & -                                 \\ \hline
\textbf{\%Unanswerable}              & \begin{tabular}[c]{@{}c@{}}5542\\ 27\%\end{tabular}   & \begin{tabular}[c]{@{}c@{}}1382\\ 32\%\end{tabular} & -                                 \\ \hline
\end{tabular}
\end{table}

\subsubsection{Evaluation metric}
The test set is confidential and not to disclose any details to researchers. All experimental results need evaluating by an online system in EvalAI\footnote{{https://evalai.cloudcv.org/web/challenges/challenge-page/523/overview}}. Specifically, in VizWiz Challenge, the Answerability task is recommended to evaluate by average precision evaluation metric which computes the weighted mean of precisions under a precision-recall curve in Equation~\ref{eq:ap}.

\begin{equation}
\label{eq:ap}
    AP = \sum_n{\left(R_n - R_{n-1} \right)\ P_n}
\end{equation}
where $R_n$ and $P_n$ are the precision and recall at the $n$-th threshold.

\subsubsection{Experimental settings}
In our architecture, we take advantage of pre-trained models in Vision and Text Transformer to extract visual and textual features. Besides, we also use fully-connected layers to normalize and reduce the dimension of features. All details of our implementation are presented in Table~\ref{tab:setting} to reproduce our model. 

\begin{table}[ht!]
\centering
\caption{Detail of experimental settings}
\label{tab:setting}
\begin{tabular}{lc}
\hline
\multicolumn{1}{c}{\textbf{Components}} & \textbf{Value}                                                                                           \\ \hline
\textbf{Vision Transformer}             & {B\_16\_imagenet1k}.                                                                               \\ \hline
\textbf{BERT}                           & bert-base-uncased                                                                                        \\ \hline
\textbf{Full-connected Layer}           & \begin{tabular}[c]{@{}c@{}}cat: 768 - 512 - 1024 - 512 - 1\\ mul: 768 - 512 - 512 - 512 - 1\end{tabular} \\ \hline
\textbf{Vector operation}               & \multicolumn{1}{l}{multiplication, concatenation}                                                        \\ \hline
\textbf{Optimizer}                      & \multicolumn{1}{l}{AdamW(lr = 3e-5, eps = 1e-8)}                                                         \\ \hline
\end{tabular}
\end{table}

\subsection{Results}
However, after two years of this challenge, it exists no publication on this task despite its necessity and importance. Therefore, in these results, we mention two strong baselines that include VWTest\footnote{https://eval.ai/web/challenges/challenge-page/523/leaderboard/1462} and BERT-RG-Regression~\cite{bertrg}. Firstly, VWTest obtains the best performance in the VizWiz Contest 2020. Despite its no publication,  VWTest is trustworthy enough to be considered as a reference. Secondly, BERT-RG~\cite{bertrg} recently obtains state-of-the-art results in the Yes/No question type. The strength of BERT-RG is to combine both residual and global features from ResNet and VGG. Two of them are typical in Convolutional approaches remaining dominant in image understanding. This characteristic is suitable for our work to compare Vision Transformer against traditional methods. We reproduce BERT-RG~\cite{bertrg} whose classifier is changed into a regression via linear layer as BERT-RG-Regression. Therefore, we consider BERT-RG-Regression as a competitive baseline in our comparison.  
\begin{table}[ht!]
\centering
\caption{The comparison results against strong baselines}
\label{tab:result}
\begin{tabular}{lcc}
\hline
\multicolumn{1}{c}{\textbf{Model}}                                                    & \textbf{Average Precision} & \multicolumn{1}{l}{\textbf{F1-score}} \\ \hline
\textbf{VWTest}                                                                       & 26.84                      & 42.32                                 \\ \hline
\textbf{\begin{tabular}[c]{@{}l@{}}BERT-RG~\cite{bertrg} \\ Regression\end{tabular}}                & 52.22                      & 41.85                                 \\ \hline\hline

\textbf{\begin{tabular}[c]{@{}l@{}}VT-Transformer\\ (Our model) \end{tabular}} & \textbf{76.96}               & \textbf{67.26}                          \\ \hline
\end{tabular}
\end{table}

The detail of the comparison between our model and strong baselines is shown in Table~\ref{tab:result}. In both Average Precision and F1-score, our model outperforms the strong baselines. Apparently, our consideration in the regression task is more suitable than traditional approaches. In average precision, the enhancement of regression against classification in the problem statement proves clearly. In F1-score, although results depend on annotations of VQA task for unanswerable class, our model also obtains a significant improvement by 25\%    

\subsection{Ablation Study}
In this part, we also conduct ablation studies of components in our architecture. Firstly, we would like to reveal the strength of Vision Transformer into Image Embedding. In this component, we compare Vision Transformer against two powerful image classification models that consist of ResNet~\cite{resnet} and VGG~\cite{vgg}. In experiments, we only evaluate the pre-trained models among them. Specifically, we use pre-trained parameters of ResNet-152 and VGG-16 in the Pytorch library. Besides, we also compare the effect of vector operations consisting of multiplication and concatenation. The detail of our results in this ablation study is presented in Table~\ref{tab:imgEmbed}. 
\begin{table}[ht!]
\centering
\caption{Ablation studies on Image Embedding modules}
\label{tab:imgEmbed}
\begin{tabular}{cccc}
\hline
\textbf{\begin{tabular}[c]{@{}c@{}}Vector \\ Operation\end{tabular}} & \textbf{Model}                              & \textbf{\begin{tabular}[c]{@{}c@{}}Average \\ Precision\end{tabular}} & \multicolumn{1}{l}{\textbf{F1-score}} \\ \hline
\multirow{3}{*}{\textbf{MUL}} & \textbf{ResNet152} & 63.13 & 63.59\\ \cline{2-4}& 
\textbf{VGG16} & 70.75& 66.82\\ \cline{2-4}&
\multicolumn{1}{l}{\textbf{VT-Transformer}} & \textbf{76.96} & \textbf{67.26} \\\hline

\multirow{3}{*}{\textbf{CAT}}   & \textbf{ResNet152} & 67.74& 65.16 \\ \cline{2-4} & 
\textbf{VGG16} & 71.57& 63.62\\ \cline{2-4}& \multicolumn{1}{l}{\textbf{VT-Transformer}} & \textbf{74.91}& \textbf{66.70}                          \\ \hline
\end{tabular}
\end{table}
In this comparison, our model, VT-Transformer, outperforms ResNet and VGG by approximately 10\% on Average Precision and 3\% on F1-score. It proves that Vision Transformer works well on Answerability instead of traditional image models. Transformer architecture brings the success of feature extraction in both vision and text. In somehow, consistent architectures in image and question embedding lead to the effectiveness in optimization. Besides, we observe that multiplication is better than concatenation in our architecture.

Secondly, we also reveal the effects of pre-trained parameters in deploying our Answerability system. In this comparison, we only consider multiplication as a vector operation to combine visual and textual features. VT-Transformer inited by pre-trained parameters works well in Answerability. This evaluation is presented in Table~\ref{tab:pretrained}. Successes of pre-trained systems with the fine-tuning mechanism come from their strong optimization in huge datasets. 

\begin{table}[ht!]
\centering
\caption{Effects of pre-trained parameters in VT-Transformer}
\label{tab:pretrained}
\begin{tabular}{lcl}
\hline
\multicolumn{1}{c}{\textbf{Models}} & \textbf{\begin{tabular}[c]{@{}c@{}}Average \\ Precision\end{tabular}} & \textbf{F1-score} \\ \hline
\textbf{w/o pretrained parameter}   & 58.95                        & 57.01               \\ \hline
\textbf{with pretrained parameter}  & \textbf{76.96}             & \textbf{67.26}    \\ \hline
\end{tabular}
\end{table}

Finally, we also present the impacts of different sizes on the Vision Transformer. The details of comparison and implementation are shown in Table~\ref{tab:vtsize}. Settings of B-16 and VT-16 include Transformer blocks, hidden size, fully-connected layer's dim, and the number of heads. Furthermore, we only consider both of them without pre-trained parameters. The completed size of the VT-Transformer is approximately 2400MB and 460MB corresponding to B-16 and VT-16. The bigger model is, the more capacity it has. Although B-16 is better than VT-16, a smaller version, VT-16, also gets a high performance against two baselines in the VizWiz-2020 dataset. 

\begin{table}[ht!]
\centering
\caption{A comparison of Vision Transformer architectures}
\label{tab:vtsize}
\begin{tabular}{ccc}
\hline
\textbf{Model} & \textbf{\begin{tabular}[c]{@{}c@{}}AveragePrecision\end{tabular}} & \textbf{F1-score} \\ \hline
\textbf{\begin{tabular}[c]{@{}c@{}}B-16\\ 12 - 768 - 3072 - 12\end{tabular}} & \textbf{58.95}                                                         & \textbf{57.01}      \\ \hline
\textbf{\begin{tabular}[c]{@{}c@{}}VT-16\\ 3 - 512 - 512 - 8\end{tabular}}   & 56.38                                                                & 47.87             \\ \hline
\end{tabular}
\end{table}
\section{Conclusion}
\label{sec:conclusion}
In this paper, we propose a Vision-Text Transformer model to overcome the challenges of the Answerability task. Our model takes advantage of pre-trained models and Transformer architecture. By formulating Answerability in the regression task, we propose a novel approach that integrates both Vision and Text Transformer to understand multi-modal data. Through specific experiments and ablation studies, our model outperforms competitive baselines in VizWiz 2020 dataset. 

\section*{Acknowledgment}

This work was supported by JSPS Kakenhi Grant Number 20H04295, 20K2046, and 20K20625. The research also was supported in part by the Asian Office of Aerospace R\&D (AOARD), Air Force Office of Scientific Research (Grant no. FA2386-19-1-4041)


\bibliographystyle{IEEEbib}
\bibliography{refs}

\end{document}